\documentclass{article}

\usepackage{microtype}
\usepackage{graphicx, float}
\usepackage{subcaption}
\usepackage{booktabs} 

\usepackage{hyperref}

\usepackage[accepted]{icml2023MFPL}

\usepackage{amsmath}
\usepackage{amssymb}
\usepackage{mathtools}
\usepackage{amsthm}

\usepackage[capitalize,noabbrev]{cleveref}

\theoremstyle{plain}

\theoremstyle{definition}

\theoremstyle{remark}

\usepackage[textsize=tiny]{todonotes}

\usepackage[utf8]{inputenc}
\usepackage{amsfonts}

\usepackage{tikz}
\usepackage[inline,shortlabels]{enumitem}
\usepackage{bm}
\setitemize{wide,nosep,topsep=0pt,labelindent=10pt}
\setenumerate{wide,nosep,topsep=0pt,labelindent=10pt}
\setdescription{wide,nosep,topsep=0pt,labelindent=10pt}

\newcommand{\vJ}{\bm J} 
\newcommand{\w}{{\bm w}} 
\newcommand{\nO}{K} 

\usepackage{amsmath}

\newcommand{\Expect}{\mathbb E}

\begin{document}

\twocolumn[

\icmltitle{Fairness in Preference-based Reinforcement Learning} 

\icmlsetsymbol{equal}{*}

\begin{icmlauthorlist}
\icmlauthor{Umer Siddique}{utsa}
\icmlauthor{Abhinav Sinha}{utsa}
\icmlauthor{Yongcan Cao}{utsa}
\end{icmlauthorlist}

\icmlaffiliation{utsa}{Unmanned Systems Lab, Department of Electrical and Computer Engineering, The University of Texas at San Antonio, San Antonio, TX, 78249, USA}

\icmlcorrespondingauthor{Umer Siddique}{muhammadumer.siddique@my.utsa.edu}

\icmlkeywords{Machine Learning, ICML}

\vskip 0.3in
]



\printAffiliationsAndNotice{}  

\begin{abstract}
In this paper, we address the issue of fairness in preference-based reinforcement learning (PbRL) in the presence of multiple objectives. The main objective is to design control policies that can optimize multiple objectives while treating each objective fairly. Toward this objective, we design a new fairness-induced preference-based reinforcement learning or FPbRL. The main idea of FPbRL is to learn vector reward functions associated with multiple objectives via new \textit{welfare-based} preferences rather than \textit{reward-based} preference in PbRL, coupled with policy learning via maximizing a generalized Gini welfare function. Finally, we provide experiment studies on three different environments to show that the proposed FPbRL approach can achieve both efficiency and equity for learning effective and fair policies.

\end{abstract}

\section{Introduction} \label{sec:intro}
The broad application of reinforcement learning (RL) faces a significant challenge, namely, the design of appropriate reward functions that align with specific mission objectives in given environments. To mitigate this challenge, preference-based RL (PbRL) (see, for example, \cite{christiano2017deep}) has emerged as a promising paradigm, leveraging human feedback to eliminate the need for manual reward function design. However, real-world missions often entail multiple objectives and the consideration of preferences among diverse users, necessitating a balanced approach. Existing PbRL methods primarily focus on maximizing a single performance metric, neglecting the crucial aspect of equity or fairness, e.g., \cite{stiennon2020summarize, wu2021recursively, lee2021pebble}. Consequently, the lack of fairness considerations poses a barrier to the widespread deployment of PbRL for systems affecting multiple end-users when it is critical to address fairness among these users.

To address this critical gap, the development of methods enabling fairness in PbRL becomes imperative. While recent advancements have explored fairness in RL, albeit not within the PbRL framework, notable contributions in, e.g., \cite{Weng19, SiddiqueWengZimmer20, fan2022welfare}, have employed welfare functions to ensure fairness in the single-agent RL setting. Furthermore, the work in~\cite{zimmer2021learning} considered fairness in a multi-agent RL setting.

This paper proposes an approach that builds upon existing studies on fairness, focusing on a PbRL setting. In particular, rather than relying on known ground truth rewards, our method involves learning fair policies by incorporating fairness directly into the PbRL paradigm, thereby eliminating the need for hand-crafted reward functions. 
By doing so, we aim to address fairness in PbRL without compromising on its advantages.

\paragraph{Contributions.} In this paper, we present a novel approach that addresses fairness in PbRL. 
{Our proposed method introduces a novel technique to learn vector rewards associated with multiple objectives by leveraging welfare-based preferences rather than reward-based preferences in~\cite{christiano2017deep}. Hence, the proposed approach provides new insights and techniques to address fairness in PbRL. 
We validate the effectiveness of our approach through comprehensive experiments conducted in three real-world domains.}
The proposed approach is expected to provide solutions for RL problems when reward functions are absent, or it is too costly to design them. 

\section{Related Work}
The concept of having equity and fairness, especially in real-world missions with multiple objectives and diverse users, is imperative. Such a concept has also been given careful consideration in many domains, including economics \cite{moulin2004fair}, political philosophy \cite{rawls2020theory}, applied mathematics~\citep{bramsFairDivisionCakeCutting1996} operations research~\cite{BauerleOtt11}, and theoretical computer science~\cite{ogryczak2014fair}. 
Fairness considerations have been incorporated into classic continuous and combinatorial optimization problems in scenarios where the underlying model was assumed to be fully known, and learning might not be necessary~\citep{neidhardtDataFusionOptimal2008, ogryczak2013compromise, NguyenWeng17,busa2017multi,
agarwal2018reductions}. Such methods include linear programming and other model-based algorithms that consider the feedback effects and dynamic impacts in decision-making processes, allowing for the development of fair policies that adapt to changing circumstances. While such methods yielded satisfactory results, they cannot be directly used if the underlying model is unknown or too complex to be modeled.

The study of fairness in RL, especially within a model-free paradigm, has gained significant attention in recent years, with notable contributions shedding light on various aspects of this emerging field. Initial work by \cite{jabbari2017fairness} laid the foundation by focusing on scalar rewards, paving the way for further advancements. Researchers have pursued diverse directions to incorporate fairness into RL frameworks. 
\citep{WenBastaniTopcu21} explored fairness constraints as a means to reduce discrimination, while the work of \cite{Jiang2019, zimmer2021learning, ju2023achieving} delved into achieving fairness among agents. The work of \citet{SiddiqueWengZimmer20} introduced a novel fair optimization problem within the context of multi-objective RL, enabling modifications to the existing deep RL algorithms to ensure fair solutions. \citet{chen2021bringing} extended the scope by incorporating fairness into actor-critic RL algorithms, optimizing general fairness utility functions for real-world network optimization problems. 
The work of \citet{zimmer2021learning}, on the other hand, focused on fairness in decentralized cooperative multi-agent settings, developing a framework involving self-oriented and team-oriented networks concurrently optimized using a policy gradient algorithm. Notably, the work in \citet{ju2023achieving} introduced online convex optimization methods as a means to learn fairness with respect to agents.

Despite the significant successes achieved in the field of deep RL, these methods heavily rely on the availability of known reward functions.
However, in many real-world problems, the task of defining a reward function is often challenging and sometimes even infeasible.
To address this limitation, PbRL has emerged as an active area of research~\cite{christiano2017deep}.
Within PbRL, different settings have been explored, depending on whether the involvement of humans is direct or if simulated human preferences are derived from the ground truth rewards.
In the context of PbRL, the standard approach typically revolves around maximizing a single criterion, such as a reward, which is inferred from the preferences~\cite{stiennon2020summarize, lee2021pebble, wu2021recursively}. 
However, it is clear that focusing exclusively on maximizing rewards falls short of assuring fairness across various objectives.
Our approach, which is consistent with the fundamental concepts of preference-based learning, digs into the investigation of learning fair policies in the context of PbRL.

\section{Preliminaries}

\subsection{Preference-based RL (PbRL)}
We consider a Markov Decision process without reward (MDP\textbackslash R) augmented with preferences, which is a tuple of the form $(\mathcal{S}, \mathcal{A}, T, \rho, \gamma)$, where $\mathcal{S}$ is the set of states, $\mathcal{A}$ is the set of possible actions, $T: \mathcal{S} \times \mathcal{A} \times \mathcal{S} \to [0, 1]$ is a state transition probability function specifying the probability $p(s' \mid s, a)$ of reaching state $s' \in \mathcal{S}$ after taking action $a$ in state $s$,  
$\gamma$ is a discount factor, and $\rho: \mathcal{S} \to [0, 1]$ specifies the initial state distribution. The learning agent interacts with the environment through rollout trajectories, where a length-$k$ trajectory segment takes the form $(s_1, a_1, s_1, a_1, \ldots, s_k, a_k)$. A \textit{policy} $\pi$ is a function that maps states to actions, such that $\pi(a \mid s)$ is the probability of taking action $a \in \mathcal{A}$ in state $s \in \mathcal{S}$.

PbRL is an approach to learning policies without rewards in which humans are asked to compare pairs of trajectories and give relative preferences between them~\cite{christiano2017deep}. More specifically, in PbRL, a human is asked to compare a pair of length-$k$ trajectory segments $\sigma^1 = (s^1_1,a^1_1,s^2_1,a^2_1,\ldots,s_k^1,a_k^1)$ and $\sigma^2 = (s^2_1,a^2_1,s^2_2,a^2_2,\ldots,s_k^2,a_k^2)$, where $\sigma^1 \succ \sigma^2$ indicates that the user preferred $\sigma^1$ over $\sigma^2$.
Owing to the unavailability of the reward function, many PbRL algorithms learn an estimated reward function model, $\hat{r}(\cdot,\cdot): \mathcal{S} \times \mathcal{A} \to \mathbb{R}$. The reward estimate $\hat{r}(\cdot,\cdot)$ can be viewed as an underlying latent factor
explaining human preferences. In particular, it is often assumed that the human’s probability of preferring a segment
$\sigma^1$ over $\sigma^2$ is given by the Bradley-Terry model~\cite{christiano2017deep},
\vspace{-.24em}
\begin{equation}\label{eq:prob}
    P(\sigma^1 \succ \sigma^2 \mid \hat{r}) = \frac{e^{\hat{R}(\sigma^1)}}{e^{\hat{R}(\sigma^1)}+e^{\hat{R}(\sigma^2)}},
\end{equation}
where $\hat{R}(\sigma_i) := \sum_{t = 1}^{k} \gamma^{t-1} \hat{r}(s_t^i, a_t^i)$ is the estimated total discounted reward of trajectory segment $\sigma_i$, and $(s_t^i, a_t^i)$ is the $t$\textsuperscript{th} state-action pair in $\sigma_i.$
One can minimize the cross-entropy loss between the Bradley-Terry preference predictions and true human preferences, given by~\cite{christiano2017deep},
\vspace{-.24em}
\begin{align}\label{eq:loss}
    L(\hat{r}) =& - \sum_{(\sigma^1, \sigma^2, \mu) \in S}\left( \mu(1) \log P[\sigma^1 \succ \sigma^2]\right.\nonumber\\
    &\left.+ \mu(2) \log P[\sigma^2 \succ \sigma^1] \right),
\end{align}
where $\mu(i),~ i \in \{1, 2\}$ is an indicator such that $\mu(i) = 1$ when trajectory segment $\sigma^i$ is preferred, whereas $S$ is the dataset of labeled human preferences.
By optimizing $L(\hat{r})$, an estimated reward function $\hat{r}(\cdot,\cdot)$ can be obtained to help explain human preferences. 

\subsection{Notion of Fairness}
\label{sec:fairness}
The fairness concept used in previous work such as~\cite{SpeicherHeidariGrgicHlacaGummadiSinglaWellerZafar18, Weng19, SiddiqueWengZimmer20, zimmer2021learning} enforces three natural properties: \textit{efficiency}, \textit{equity}, and \textit{impartiality}. 
The concept of {efficiency}, also referred to as \textit{optimality},  implies that the solution should be optimal and Pareto dominant. {Equity} is often associated with the concept of distributive justice, as it pertains to the fairness of resource or opportunity distribution.
This property ensures that a fair solution follows the {Pigou-Dalton principle}~\cite{Moulin04}, which states that by transferring rewards from the more advantaged to the less advantaged users, the overall fairness of the solution can be improved. {Impartiality} or {equality} requires that all users be treated equally, without favoritism towards any particular user in terms of the solution's outcomes.

To operationalize this notion of fairness, the use of welfare functions is employed.
These welfare functions aggregate the utilities of all users and provide a measure of the overall desirability of a solution for the entire group.
While there exist various welfare functions, we only consider those that satisfy the three fairness properties discussed earlier. One welfare function that satisfies the aforementioned properties is the \textit{generalized Gini welfare function} \citep{Weymark81}, which is defined as follows:
\begin{align} \label{eq:ggi}
    \phi_{\bm w}(\bm u) = \sum_{i \in \mathcal{K}} \w_i \bm u_i^\uparrow \,,
\end{align}
where $\bm u \in \mathbb R^\mathcal{K}$ represents the utility vector of a size $\mathcal{K}$, $\w \in \mathbb R^\mathcal{K}$ is a fixed weight vector with positive components that strictly decrease (i.e., $\w_1 > \ldots > \w_\nO$), and $\bm u^\uparrow$ denotes the vector obtained by sorting the components of $\bm u$ in increasing order (i.e., $\bm u^\uparrow_1 \le \ldots \le \bm u^\uparrow_\nO$).
{For consistency, bold variables represent vectors/matrices.}
In essence, this function computes the summation of the weight multiplied by the sorted utility for each objective. 
The weight vector is fixed, positive, and strictly decreasing.
It is important to note that the strict decrease in weights is crucial to ensure a fair and Pareto optimal, as well as an equitable solution.

\section{Approach}
In order to account for the impact of an agent's actions on multiple objectives, i.e., users in the notion of fairness in ~\Cref{sec:fairness}, we extend previous RL formulations by redefining the estimated reward function as a vector function, denoted as $\bm{\hat{r}}: \mathcal{S} \times \mathcal{A} \rightarrow \mathbb{R}^\mathcal{K}$, where $\mathcal{K}$ denotes the number of objectives. This vector function captures the rewards associated with all objectives, acknowledging the multi-objective nature of the problem at hand. Note that this is different from the scalar reward function $\hat{r}$ in PbRL~\cite{christiano2017deep}. To formalize the fair policy optimization problem, we integrate the welfare function $\phi_{\w}$ into our objective function. Consequently, the goal is to find a policy that generates a fair distribution of rewards over $\mathcal{\nO}$ objectives given by
\begin{align}\label{eq:ggi pb}
\max_{\pi_{\bm \theta}} \phi_{\bm w}(\vJ(\pi_{\bm \theta})),
\end{align}
where $\pi_{\bm \theta}$ represents a policy parameterized by $\bm \theta$, $\phi_{\bm w}$ denotes a welfare function with fixed weights that requires optimization, and $\vJ(\pi_{\bm \theta})$ represents the vectorial objective function that yields the utilities (i.e., $\bm u$) for all users. It is also worth noting that the chosen welfare function, such as the generalized Gini welfare function, is concave. As a result, the optimization problem presented in  \eqref{eq:ggi pb} can be characterized as a convex optimization problem. This convexity property facilitates the exploration of effective solution methods for achieving equitable policies in model-free RL settings.

Note that optimizing the welfare function defined in~\eqref{eq:ggi} is an effective way to address fairness because the weights $\w$ are selected such that a higher weight will be assigned for objectives with lower utility values, which will ensure that all objectives are treated fairly than the cases when the weights are assigned without considering the utility values. 


Our procedure to optimize the welfare function is an iterative process that integrates the policy update step and reward update step (via the collection of more preferences for reward function estimation). Since the reward function estimation is non-stationary, we focus on policy gradient methods. 
As a state-of-the-art policy gradient method, we adopt the Proximal Policy Optimization (PPO) algorithm \citep{SchulmanWolskiDhariwalRadfordKlimov17} for policy optimization and compute the advantage function via
\begin{align}
    {\bm A}_{\pi_{\bm \theta}}(s_t,a_t) = \sum_{t} (\gamma \lambda)^{t-1} \delta_t \,,
\end{align}
where $\delta_t$ is determined by the expression ${\bm{\hat{r}}_t} + \gamma {\bm V}_{\bm \theta}(s_{t+1}) - {\bm V}_{\bm \theta}(s_t)$, with $\bm{\hat{r}}_t$ representing the estimated rewards, and ${\bm V}_{\bm \theta}(s_t)$ denoting the value function associated with state $s_t$. In PPO, the objective function $\vJ(\bm \theta)$ is designed to limit policy changes after an update, that is,
\begin{align} \label{eq:ppoobjec}
& \Expect_{s \sim {\bm d}_{\pi}, a \sim \pi_{\bm \theta}(\cdot|s)} \left[ \min(\rho_{\bm \theta} {\bm A}_{\pi_{\bm \theta}}(s,a), \bar{\rho}_{\bm \theta} {\bm A}_{\pi_{\bm \theta}}(s,a)) \right] \,,
\end{align}
where $\rho_{\bm \theta} = \dfrac{\pi_{\bm \theta}(a|s)}{\pi_{\bm b}(a|s)}$, $\bar{\rho}_{\bm \theta} = \text{clip}(\rho_{\bm \theta}, 1 - \epsilon, 1 + \epsilon)$, $\pi_{\bm b}$ represents the policy generating the transitions, and $\epsilon$ is a hyperparameter controlling the constraint. To compute the gradient for $\vJ(\bm \theta)$, we have \begin{align} \label{eq: mo objec}
\nabla_{\bm\theta} \phi_{\w} (\vJ(\pi_{\bm\theta})) =& \nabla_{\vJ(\pi_{\bm\theta})} \phi_{\w} (\vJ(\pi_{\bm\theta})) \cdot \nabla_{\bm\theta} \vJ(\pi_{\bm\theta}) \\
=& {\bm w}_\sigma^\intercal \nabla_{\bm\theta} \vJ(\pi_{\bm\theta}),
\end{align}
where $\nabla_{\bm\theta} \vJ(\pi_{\bm\theta})$ is a $\mathcal{K}\times \mathcal{N}$ matrix representing the classic policy gradient over the $\mathcal{K}$ objectives, ${\bm w}_\sigma$ is a vector sorted based on the values of $\vJ(\pi_{\bm\theta})$, and $\mathcal{N}$ denotes the number of policy parameters.  

For reward estimation function update, we ask a human (or a similar mechanism like a synthetic human) to provide preferences for the segments collected by the policy, establishing or expanding the dataset for preferences. The vector function $\bm{\hat{r}}$ is learned via minimizing the loss function~\eqref{eq:loss} with a modified preference probability given by
\begin{equation}\label{eq:mo prob}
    P(\sigma^1 \succ \sigma^2 \mid \bm{\hat{r}}) = \frac{e^{\hat{ R}(\sigma^1)}}{e^{\hat{ R}(\sigma^1)}+e^{\hat{ R}(\sigma^2)}},
\end{equation}
where $\hat{ R}(\sigma_i) := \phi_{\w}(\sum_{t = 1}^{k} \gamma^{t-1} \bm{\hat{r}}(s_t^i, a_t^i))$. 
This formulation applies the welfare function $\phi_{\bm w}$ to the discounted cumulative vector rewards, resulting in a scalarized $\hat{ R}(\sigma_i)$.
This scalarized value is then utilized to compute $P(\sigma^1 \succ \sigma^2 \mid \bm{\hat{r}})$. 
It is important to note that the key distinction between our proposed approach and PbRL in~\cite{christiano2017deep} lies in the utilization of the welfare function to determine preferences, as opposed to relying on segment rewards as done in~\cite{christiano2017deep}.

\section{Experimental Results}
To demonstrate the robustness and practicality of our method, we meticulously design and conduct three experiments. Each experiment showcases a unique scenario where fairness plays a pivotal role in RL outcomes. Moreover, at present, our primary emphasis is directed toward investigating synthetic human preferences owing to their convenient acquisition process and their appropriateness for testing objectives. Nonetheless, it is essential to note that our proposed approach is readily applicable in situations that involve human-in-the-loop interactions. Through rigorous analysis and evaluation, we assess the performance of our approach, both in terms of achieving fairness objectives and maintaining desirable learning outcomes in a model-free setting. We assign weights $\bm{w}_i = \frac{1}{2^i}, i = 0,...,\mathcal{K}-1$, and to ensure the reproducibility of the results, and average the results over $5+$ runs with different seeds to provide reliable evidence of our method's effectiveness. 
All algorithm hyperparameters were optimized using the open-source Lightweight HyperParameter Optimizer (LHPO)~\cite{zimmer2018phd}.

\begin{figure*}[t]
    \centering
     \begin{subfigure}[t]{0.32\linewidth}
         \centering
        \includegraphics[width=\linewidth]{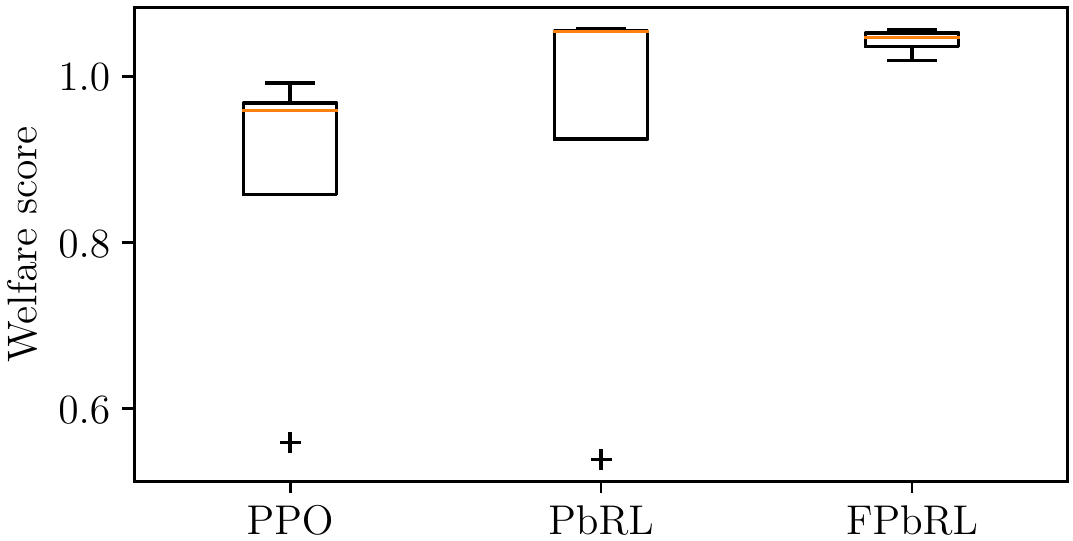}
         \caption{Welfare scores.}
         \label{fig:spec_box}
     \end{subfigure}
     \begin{subfigure}[t]{0.32\linewidth}
         \centering
         \includegraphics[width=\linewidth]{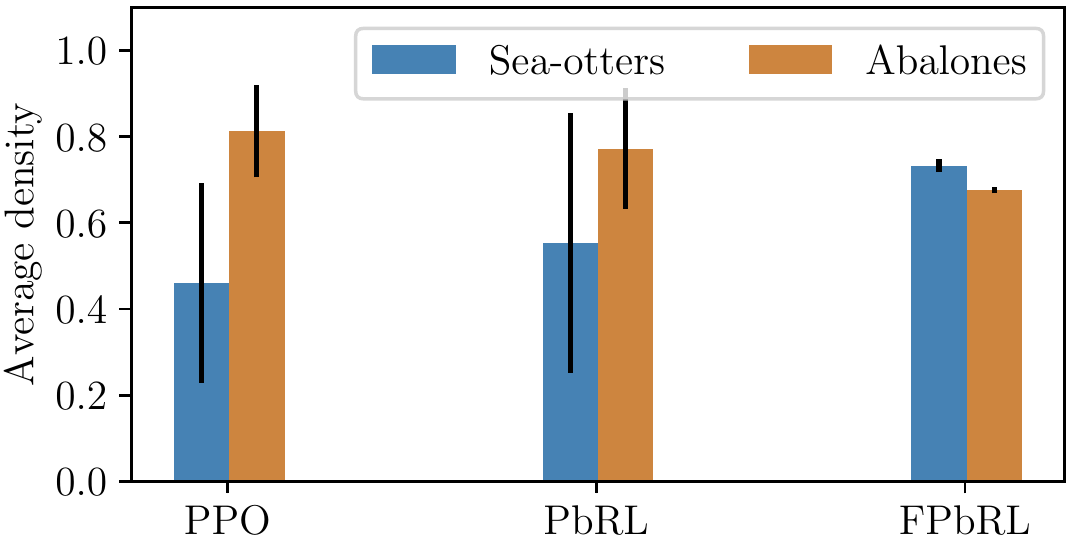}
        \caption{Individual densities.}
        \label{fig:spec_bar}
     \end{subfigure}
     \begin{subfigure}[t]{0.32\linewidth}
         \centering
         \includegraphics[width=\linewidth, height=0.5\linewidth]{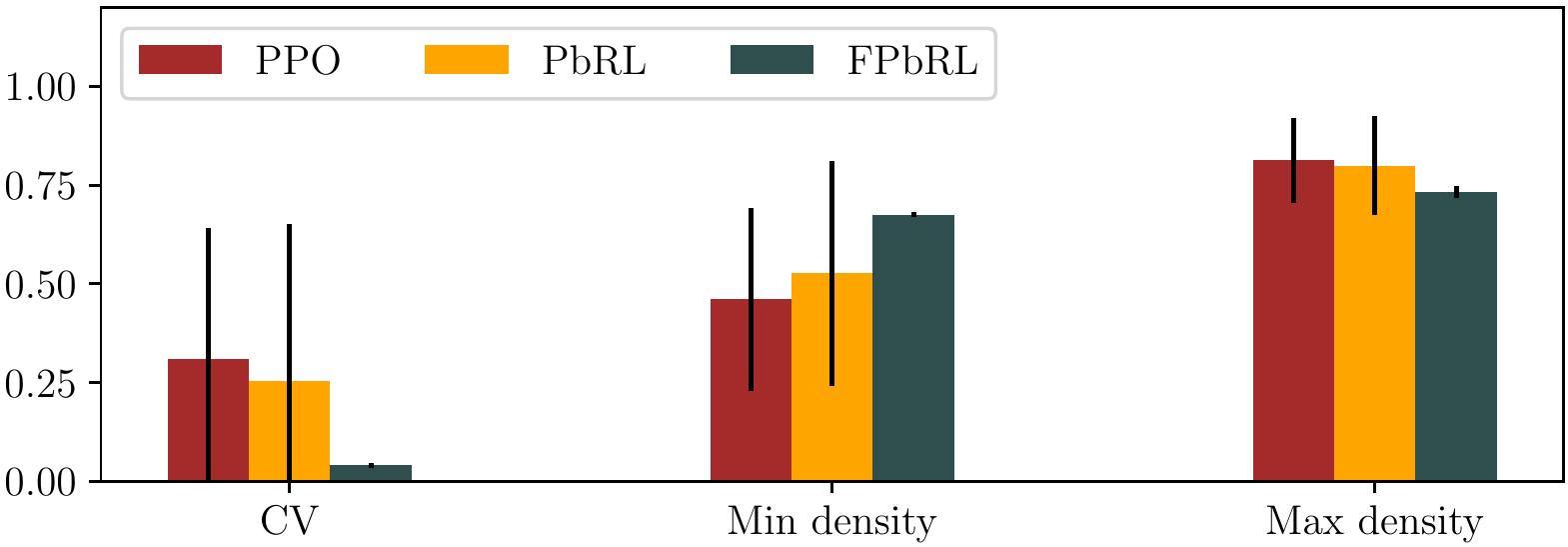}
        \caption{CV, min and max densities.}
        \label{fig:spec_cv}
     \end{subfigure}
    \caption{Performances of PPO, PbRL, FPbRL in the species conservation problem.}
    \label{fig:algos_sc}
\end{figure*}

\subsection{Species Conservation}
Species conservation is a critical domain in the field of ecology, particularly when dealing with the preservation of multiple interacting endangered species. Here, we tackle the challenge of incorporating fairness considerations into the conservation efforts of two specific species: sea otters and their prey, the northern abalone. The sea otter and northern abalone populations face a delicate balance as sea otters consume abalones, both of which are currently endangered. To navigate this complex conservation problem, we adopt the setting proposed in \citet{chades2012setting} and tailor it to address the fairness aspects of this ecosystem. In our conservation problem, the state is defined by the current population numbers of both species. To influence the system, we have five distinct actions at our disposal: introducing sea otters, enforcing antipoaching measures, controlling sea otter populations, implementing a combination of half-antipoaching and half-controlled sea otters, or taking no action.  Each action has significant implications, as introducing sea otters is necessary for balancing the abalone population, but if not carefully managed, it can inadvertently drive the abalone species to extinction. Similarly, neglecting any of the other managerial actions would result in the extinction of one of the species, highlighting the importance of a comprehensive approach in terms of equity and fairness. The transition function in this conservation problem incorporates population growth models for both species, accounting for factors such as poaching and oil spills. Through this framework, we strive to optimize not just a single objective but the population densities of both species, thereby dealing with a multidimensional problem where two objectives, sea otter and abalone population densities, need to be simultaneously optimized, leading to $\mathcal{K}=2$.

In this domain, our primary objective is to assess the effectiveness of our proposed method in optimizing the welfare function, denoted as $\phi_{\w}$. To evaluate this, we conduct a comparative analysis of welfare scores between three approaches: PPO, PbRL, and our proposed FPbRL method within this domain. To compute the welfare scores, we employ trained agents and evaluate their performance across 100 trajectories within the given environment. The empirical average vector returns of these trajectories serve as the basis for deriving the welfare score by applying the function $\phi_{\w}$. The distribution of welfare scores for PPO, PbRL, and FPbRL is shown in \Cref{fig:spec_box}. Our results reveal that FPbRL achieves the highest welfare score, thereby demonstrating its ability to identify fairer solutions compared to PPO and the standard PbRL method. However, recognizing that the welfare score alone may not provide a comprehensive understanding of the objective balance, we present individual density plots in \Cref{fig:spec_bar} depicting the densities of both species. These plots offer further insights into the distribution of objectives. Consistently, our findings demonstrate that FPbRL yields more balanced solutions in terms of equity, surpassing both PbRL and PPO. In addition, we introduce the Coefficient of Variation (CV) to address scenarios where demonstrating the utility of each objective becomes challenging due to a multitude of objectives. \Cref{fig:spec_cv} showcases the CV, as well as the minimum and maximum densities. Corresponding with our previous findings, our proposed FPbRL method exhibits the lowest CV, indicating reduced variation between different objectives. Moreover, our method prioritizes maximizing the minimum objective to foster a more equitable distribution of utilities.

\begin{figure*}[t]
    \centering
     \begin{subfigure}[t]{0.32\linewidth}
         \centering
        \includegraphics[width=\linewidth]{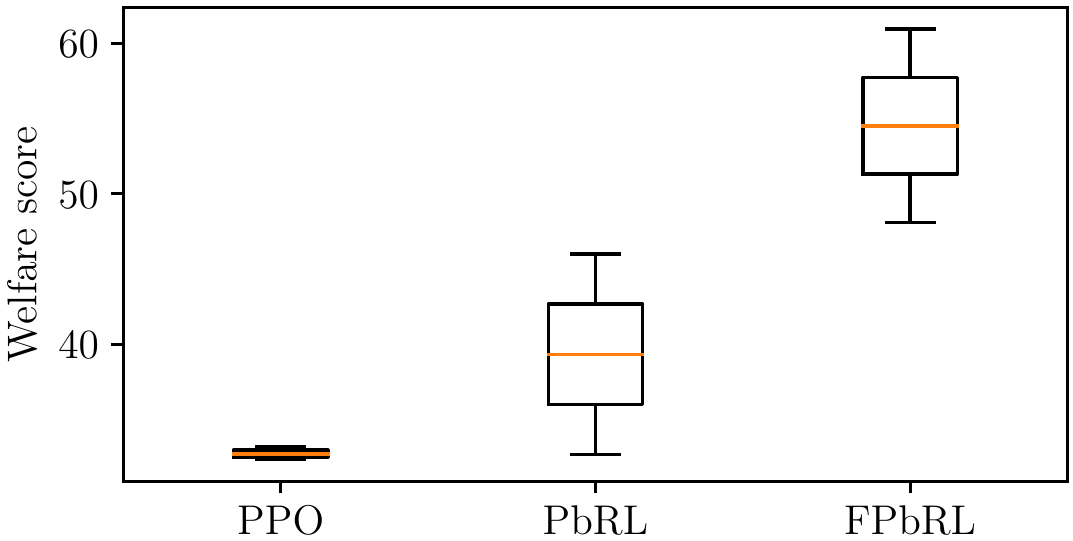}
         \caption{Welfare scores.}
         \label{fig:resource_box}
     \end{subfigure}
     \begin{subfigure}[t]{0.32\linewidth}
         \centering
         \includegraphics[width=\linewidth]{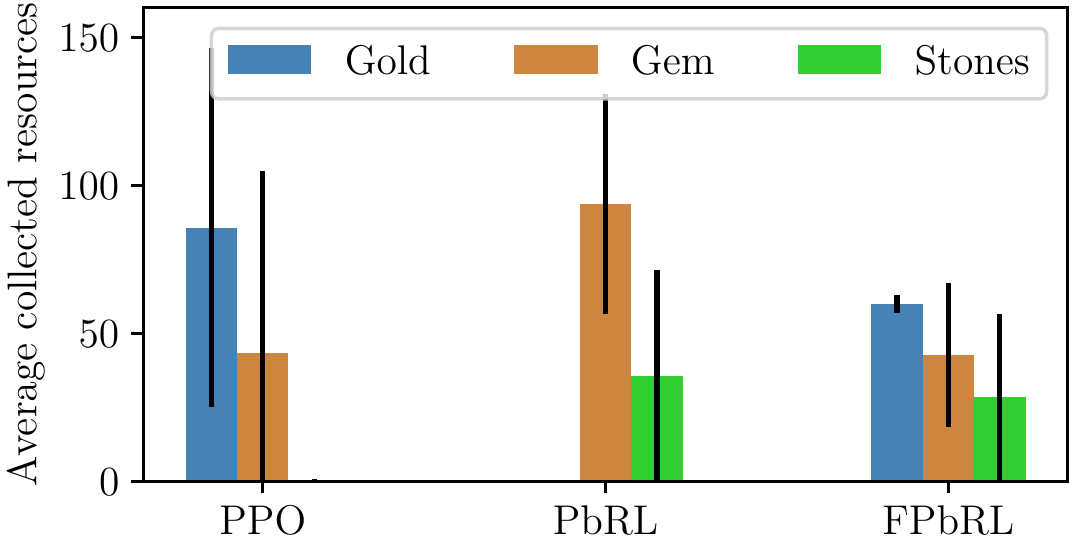}
        \caption{Individual number of resources collected.}
        \label{fig:rc_bar}
     \end{subfigure}
     \begin{subfigure}[t]{0.32\linewidth}
         \centering
         \includegraphics[width=\linewidth, height=0.5\linewidth]{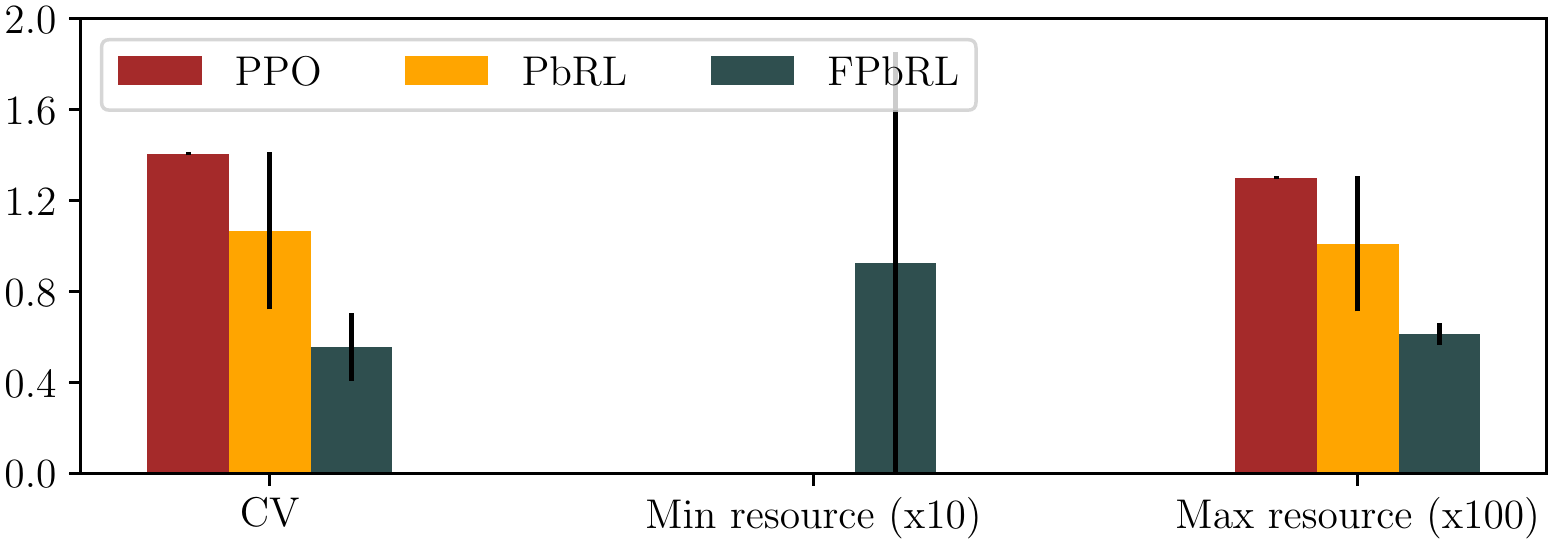}
        \caption{CV, min, and max resources collected.}
        \label{fig:resource_cv}
     \end{subfigure}
    \caption{Performances of PPO, PbRL, FPbRL in resource gathering.}
    \label{fig:algos_rc}
\end{figure*}

\subsection{Resource Gathering}
We now consider a resource-gathering environment that encompasses a $5 \times 5$ grid world, adapted from the work of \citep{barrett2008learning}. This dynamic environment poses the challenge of resource acquisition, where the agent's objective is to collect three distinct types of resources: gold, gems, and stones, thus $\mathcal{K}=3$. Within this grid world, the agent is situated at a specific position, while the resources are scattered randomly across various locations. Upon consumption of a resource, it is promptly regenerated at another random location within the grid, ensuring a continuous supply. The state representation in this environment encapsulates the agent's current position within the grid, as well as the cumulative count of each resource type collected throughout the ongoing trajectory. To navigate this complex environment, the agent is equipped with four cardinal direction actions: up, down, left, and right, enabling movement across the grid. However, to introduce an additional layer of intricacy, we assign distinct values to the resources. Gold and gems are endowed with a value of 1, symbolizing their higher significance, while stones, deemed less valuable, are assigned a value of 0.4. This deliberate assignment fosters an unbalanced distribution of resources, with two stones, one gold, and one gem, strategically placed within the grid. Amidst this resource-rich environment, the agent's ultimate goal is twofold: to maximize the accumulation of resources while concurrently maintaining a balanced distribution among the different resource types. By striking this delicate equilibrium, the agent strives to optimize its resource-gathering strategy, maximizing its overall utility and adaptability within this domain.

\begin{figure*}[t]
    \centering
     \begin{subfigure}[t]{0.32\linewidth}
         \centering
        \includegraphics[width=\linewidth]{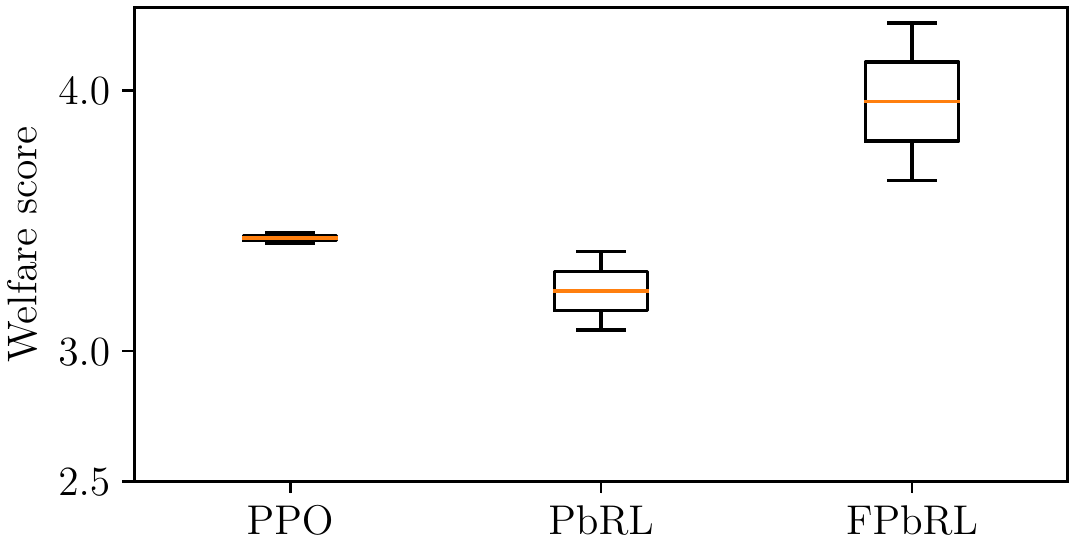}
         \caption{Welfare scores.}
         \label{fig:tl_box}
     \end{subfigure}
     \begin{subfigure}[t]{0.32\linewidth}
         \centering
         \includegraphics[width=\linewidth]{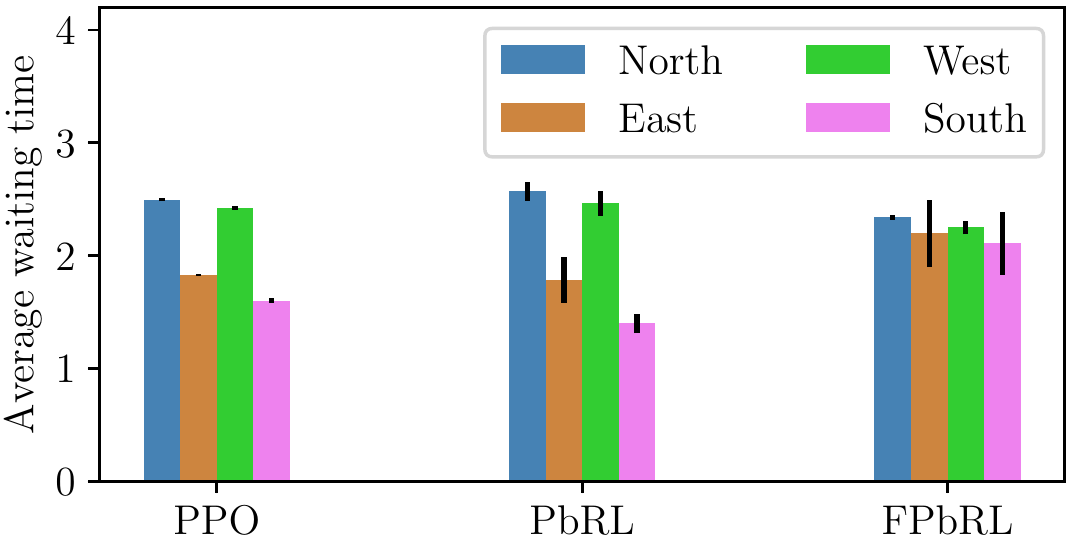}
        \caption{Individual waiting times.}
        \label{fig:tl_bar}
     \end{subfigure}
     \begin{subfigure}[t]{0.32\linewidth}
         \centering
         \includegraphics[width=\linewidth, height=0.5\linewidth]{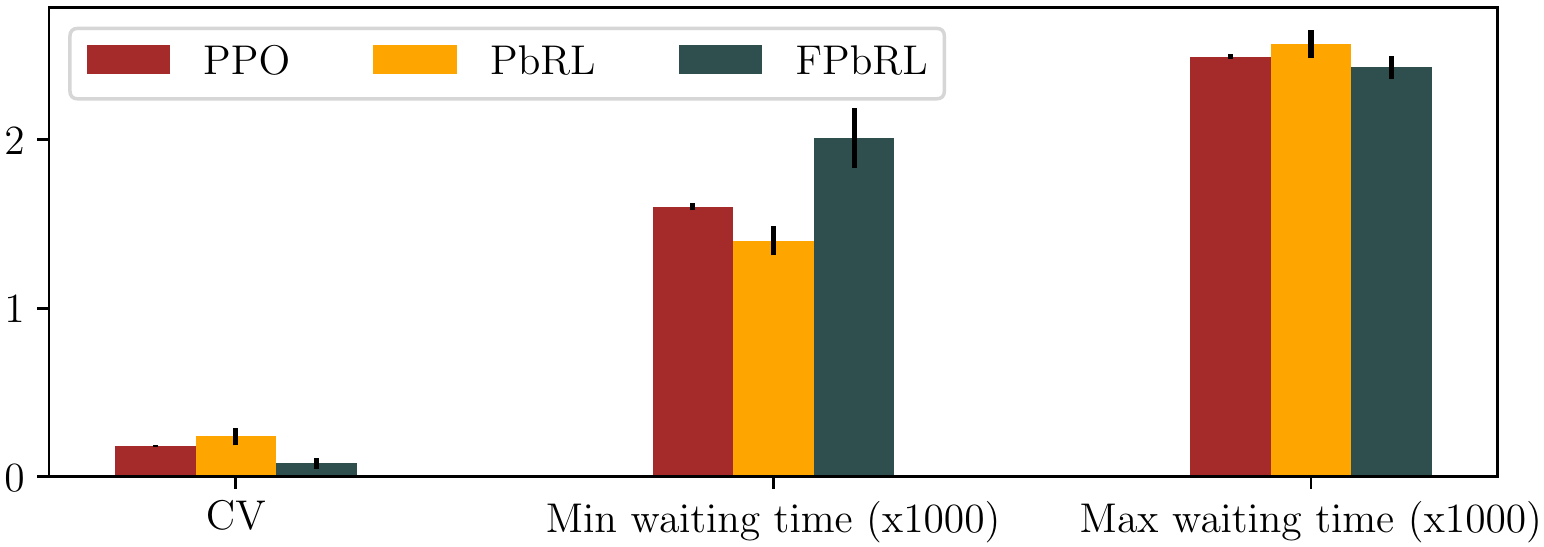}
        \caption{CV, min, and max waiting times}
        \label{fig:tl_cv}
     \end{subfigure}
    \caption{Performances of PPO, PbRL, FPbRL in traffic control.}
    \label{fig:algos_tl}
\end{figure*}

To demonstrate the efficacy of our proposed approach in maintaining a balanced distribution of resources, we conduct an analysis of welfare scores for the resource collection problem. Through this analysis, we aim to assess the fairness of different approaches and determine the extent to which our proposed method promotes equitable solutions. \Cref{fig:resource_box} presents the welfare scores computed for PPO, PbRL, and the proposed FPbRL. These scores were computed over a hundred trajectories during the testing phase. Encouragingly, our proposed method achieved the highest welfare score, signifying a fairer solution when compared to both PPO and the standard PbRL method. To gain a comprehensive understanding of the balance between objectives in resource collection, we also examine the individual number of resources collected (see \Cref{fig:rc_bar}). Once again, the results reinforce the superiority of FPbRL in producing more balanced solutions. In contrast, PbRL and PPO tend to favor the accumulation of certain resources at the expense of others, highlighting the limitations of a standard approach that solely optimizes the aggregate or weighted sum of objectives. Our proposed method, however, maintains a balanced distribution of different resources, underscoring the significance of fairness considerations in resource collection scenarios. Furthermore, \Cref{fig:resource_cv} provides additional insights into the performances of PPO, PbRL, and FPbRL by examining the CV as well as the minimum and the maximum number of collected resources. Strikingly, FPbRL outperforms the other algorithms, exhibiting the lowest CV, which indicates a more equitable distribution of objectives. Notably, only FPbRL successfully maximizes the minimum objective utility, whereas PPO and the PbRL method yield the lowest minimum objective values, reflecting a prioritization of maximizing cumulative rewards at the expense of fairness considerations.

\subsection{Traffic Control at Intersections}
To thoroughly validate the effectiveness of our proposed method, we also conduct a series of experiments in the demanding real-world domain of traffic light control. This domain presents unique challenges due to the multitude of objectives involved, making it an ideal testbed for evaluating the efficacy of our approach. To simulate a realistic traffic intersection scenario, we employed the widely-used Simulation of Urban MObility (SUMO) platform~\citep{SUMO2018}. 
Specifically, our focus is on a standard 8-lane intersection, with two lanes designated for turning (left or right, depending on the side of the road) and the remaining lanes facilitating straight driving or additional turns. Traditionally, the objective of traffic control is to optimize traffic flow by minimizing the total waiting time for all vehicles approaching the intersection. However, our approach diverges from this conventional perspective. Instead, we adopted a novel viewpoint, aiming to optimize traffic flow for each of the four distinct sides of the road. Each side of the intersection is treated as a separate objective, and our goal is to learn a controller that effectively reduces the expected waiting times for vehicles on each road segment. This multi-objective setup thus takes $\mathcal{K}=4$, reflecting the four sides of the road that need to be individually optimized. In this challenging problem, a state is defined by several key factors, including the waiting time of vehicles, the car density in the vicinity, and the current phase of the traffic light. The action space comprises four distinct options, each corresponding to a different phase change that influences the traffic flow on a specific side of the road. The transition function governing the evolution of the system is dependent on factors such as the current traffic light phase, the movement of vehicles through the intersection, and the generation of new traffic.

Similar to the previous assessments, we evaluate the efficacy of the proposed method in optimizing the welfare function. The welfare scores obtained during testing for PPO, PbRL, and FPbRL are presented in \Cref{fig:tl_box}. To improve readability, the y-axis has been scaled by a factor of 1000, with each tick representing 1000 units. It is evident that FPbRL outperforms both PPO and PbRL, achieving the highest welfare score. This noteworthy result underscores the efficacy of FPbRL in optimizing the welfare function, which is crucial for ensuring fair and equitable treatment of the diverse objectives at hand. To establish the correlation between these high welfare scores and fairer solutions, we examine the waiting times for all sides of the roads, as depicted in \Cref{fig:tl_bar}. Our proposed method, FPbRL, demonstrates a more balanced distribution of waiting times across all road segments. Although FPbRL exhibits slightly higher total waiting times, it prioritizes lanes with fewer cars, thereby preventing any single vehicle from enduring significantly prolonged waiting periods. In contrast, PPO and PbRL tend to favor lanes with higher car densities in their pursuit of minimizing total waiting times. This observation underscores the importance of fairness considerations, indicating that the attainment of fairness may sometimes come at a cost. However, the cost of fairness is not excessively high, as evidenced in the previous domains (\Cref{fig:spec_bar,fig:rc_bar}). Furthermore, we compare the performances of PPO, PbRL, and FPbRL in terms of the CV, minimum waiting time, and maximum waiting time (\Cref{fig:tl_cv}). Once again, FPbRL emerges as the top-performing algorithm, attaining the lowest CV and achieving a more balanced distribution of objectives. Notably, only FPbRL successfully maximizes the minimum objective and minimizes the maximum objective, thereby promoting equitable outcomes in the context of traffic light control.

\section{Conclusions and Future Work}
By incorporating fairness into PbRL, we developed a new fairness-induced PbRL (FPbRL) approach that can provide more equitable and socially responsible RL systems. Through our multi-experiment validation, we provided compelling evidence of the effectiveness and practicality of our approach toward its potential applications in real-world scenarios where fairness considerations are imperative. Our findings underscore the effectiveness of our proposed method, FPbRL, in optimizing the welfare function and achieving fairness in the presence of multiple objectives. A detailed investigation of other welfare functions and different impartiality properties, along with actual human feedback, could be interesting to explore in the future.



\section*{Acknowledgements}
The authors were supported in part by Army Research Lab under grant W911NF2120232, Army Research Office under grant  W911NF2110103, and Office of Naval Research under grant N000142212474. 




\bibliography{references}
\bibliographystyle{icml2023}



\end{document}